# Optimizing Monotone Functions Can Be Difficult

Benjamin Doerr[1], Thomas Jansen[2], Dirk Sudholt[3,4],
Carola Winzen[1], and Christine Zarges[5]

[1]Max-Planck-Institut für Informatik, 66123 Saarbrücken, Germany
[2]University College Cork, Cork, Ireland
[3]International Computer Science Institute, Berkeley, CA, USA
[4]University of Birmingham, Birmingham B15 2TT, UK
[5]Technische Universität Dortmund, 44221 Dortmund, Germany

**Abstract**

Extending previous analyses on function classes like linear functions, we analyze how the simple (1+1) evolutionary algorithm optimizes pseudo-Boolean functions that are strictly monotone. Contrary to what one would expect, not all of these functions are easy to optimize. The choice of the constant $c$ in the mutation probability $p(n) = c/n$ can make a decisive difference.

We show that if $c < 1$, then the (1+1) EA finds the optimum of every such function in $\Theta(n \log n)$ iterations. For $c = 1$, we can still prove an upper bound of $O(n^{3/2})$. However, for $c > 33$, we present a strictly monotone function such that the (1+1) EA with overwhelming probability does not find the optimum within $2^{\Omega(n)}$ iterations. This is the first time that we observe that a constant factor change of the mutation probability changes the run-time by more than constant factors.

## 1 Introduction

Rigorously understanding how randomized search heuristics solve optimization problems and proving guarantees for their performance remains a challenging task. The current state of the art is that we can analyze some heuristics for simple problems. Nevertheless, this gave new insight, helped to get rid of wrong beliefs, and turned correct beliefs into proven facts.

For example, it was long believed that a pseudo-Boolean function $f : \{0,1\}^n \to \mathbb{R}$ is easy to optimize if it is *unimodal*, that is, if each $x \in \{0,1\}^n$ that is not optimal has a Hamming neighbor $y$ with $f(y) > f(x)$ [Müh92]. Recall that $y$ is called a Hamming neighbor of $x$ if $x$ and $y$ differ in exactly one bit.

This belief was debunked in [DJW98]. There the unimodal long $k$-path function [HGD94] was considered and it was proven that the simple (1+1) evolutionary algorithm ((1+1) EA) with high probability does not find the optimum within $2^{\sqrt{n}}$ iterations. This classical episode shows how important it is to support an intuitive understanding of evolutionary algorithms with rigorous proofs.

It also shows that it is very difficult to identify problem classes that are easy for a particular randomized search heuristic. This, however, is needed for a successful application of such methods, because the no free lunch theorems [IT04] tell us, in simple words, that no randomized search heuristic can be superior to another if we do not restrict the problem class we are interested in.

## 1.1 Previous Work

In the following, we restrict ourselves to classes of simple pseudo-Boolean functions. We stress that the last ten years also produced a number of results on combinatorial problems [OHY]. At the same time research on classical test functions and function classes continued, spurred by the many still open problems.

We also restrict ourselves to one of the most simple randomized search heuristics, the (1+1) EA. The first rigorous results on this heuristic were given by Mühlenbein [Müh92], who determined how long it takes to find the optimum of simple test functions like $\textsc{OneMax}(x) := \sum_{i=1}^{n} x_i$, counting the number of 1-bits. Quite some time later, and with much more technical effort necessary, Droste, Jansen and Wegener [DJW02] extended the $O(n \log n)$ bound to all linear functions $f(x) := \sum_{i=1}^{n} a_i x_i$. Since it was hard to believe that such a simple result should have such a complicated proof, this work initiated a sequence of follow-up results in particular introducing drift analysis to the community [HY01, HY02] or refining it for our purposes [Jäg08, DFW10, DJW10]. However, not all promising looking function classes are easy to optimize. As laid out in the first paragraphs of this paper, already unimodal functions are difficult.

Almost all results described above were proven for the standard mutation probability $1/n$. It is easy to see from their proofs (or, in the case of linear functions, cf. [DG10]), that all results remain true for $p(n) = c/n$, where $c$ can be an arbitrary constant.

We should add that the question how to determine the right mutation probability is also far from being settled. Most theory results for simplicity take the value $p(n) = 1/n$, but it is known that this is not always optimal [JW00]. In practical applications, similarly $1/n$ is the most recommended static choice for the mutation probability [BFM97, Och02] in spite of known limitations of this choice [CS09].

## 1.2 Our Work

In this work, we regard the class of strictly monotone functions. A pseudo-Boolean function is called *strictly monotone* (or simply *monotone* in the following) if any mutation flipping at least one 0-bit into a 1-bit and no 1-bit into a 0-bit strictly increases the function value. Hence much stronger than for unimodal functions, we not only require that each non-optimal $x$ has a Hamming neighbor with better $f$-value, but we even ask that this holds for all Hamming neighbors that have an additional 1-bit.

Obviously, the class of monotone functions includes linear functions with all bit weights positive. On the other hand, each monotone function is unimodal. Contrary to the long $k$-path function there is always a short path of at most $n$ search points with increasing $f$-value connecting a search point to the optimum.

It is easy to see that monotone functions are just the ones where a simple coupon collector argument shows that random local search finds the optimum in time $O(n \log n)$. Surprisingly, we find that monotone functions are not easy to optimize for the (1+1) EA in general. Secondly, our results show that for this class of functions, the mutation probability $p(n) = c/n$, $c$ a constant, can make a crucial difference.

More precisely, we show that for $c < 1$ the (1+1) EA with mutation probability $c/n$ finds the optimum of any monotone function in time $\Theta(n \log n)$, which is best possible given previous results on linear functions. For $c = 1$, the drift argument breaks down and we have resort to an upper bound of $O(n^{3/2})$ based on a related model by Jansen [Jan07]. We currently do not know what is the full truth. As lower bound, we only have the general lower bound $\Omega(n \log n)$ for all mutation-based evolutionary algorithms.

If $c$ is sufficiently large, an unexpected change of regime happens. For $c > 33$, we show that there are monotone functions such that with overwhelming probability, the (1+1) EA does not find the optimum in exponential time. The construction of such functions heavily uses probabilistic methods. To the best of our knowledge, this is the first time that problem instances are constructed this way in the theory of evolutionary computation.

## 2 Preliminaries

We consider the maximization of a pseudo-Boolean function $f\colon \{0,1\}^n \to \mathbb{R}$ by means of a simple evolutionary algorithm, the (1+1) EA. The results can easily be adapted for minimization. In this work, $n$ always denotes the number of bits in the representation.

The (1+1) EA (Algorithm 1) maintains a population of size 1. In each generation it creates a single offspring by independently flipping each bit in the current search point with a fixed mutation probability $p(n)$. The new search point replaces the old one in case its $f$-value is not worse.

---
**Algorithm 1** (1+1) EA with mutation probability $p(n)$
---
1: **Initialization:** Choose $x \in \{0,1\}^n$ uniformly at random.
2: **repeat forever**
3:     Create $y \in \{0,1\}^n$ by copying $x$.
4:     **Mutation:** Flip each bit in $y$ independently with probability $p(n)$.
5:     **Selection: if** $f(y) \geq f(x)$ **then** $x := y$.
---

In our analyses we denote by $\mathrm{mut}(x)$ the bit string that results from a mutation of $x$. We denote as $x^+$ the search point that results from a mutation of $x$ and a subsequent selection. Formally, $y = \mathrm{mut}(x)$ and $x^+ = y$ if $f(y) \geq f(x)$ and $x^+ = x$ otherwise.

For $x = x_1 \cdots x_n$ let $Z(x)$ describe the positions of all 0-bits in $x$, $Z(x) := \{1 \leq i \leq n \mid x_i = 0\}$. By $|x|_0 = |Z(x)|$ we denote the number of 0-bits in $x$ and by $|x|_1 = n - |x|_0$ we denote the number of 1-bits. For $k \in \mathbb{N}$ let $[k] := \{1, 2, \ldots, k\}$. For a set $I = \{i_1, i_2, \ldots, i_\ell\} \subseteq [n]$ we write $x_{|I} = x_{i_1} x_{i_2} \cdots x_{i_\ell}$ for the sub-string of $x$ with the bits selected by $I$. To simplify notation we assume that any time we consider some $r \in \mathbb{R}_0^+$ but in fact need some $r' \in \mathbb{N}_0$ we assume that $r$ is silently replaced by $\lfloor r \rfloor$ or $\lceil r \rceil$ as appropriate.

We are interested in the optimization time, defined as the number of mutations until a global optimum is found. For the (1+1) EA this is an accurate measure of the actual run time. For bounds on the optimization time we use common asymptotic notation. A run time bound is called *exponential* if it is $2^{\Omega(n)}$. We also say that an event $A$ occurs *with overwhelming probability (w. o. p.)* if $1 - \Pr(A) = 2^{-\Omega(n)}$.

A function $f$ is called *linear* if it can be written as $f(x) := \sum_{i=1}^n a_i x_i$ for weights $a_1, \ldots, a_n \in \mathbb{R}$. The most simple linear function is the function $\mathrm{OneMax}(x) := \sum_{i=1}^n x_i = |x|_1$. Another intensively studied linear function is $\mathrm{BinVal}(x) := \sum_{i=1}^n 2^{n-i} x_i$. As $2^{n-i} > \sum_{j=i+1}^n 2^{n-j}$, the bit value of some bit $i$ dominates the effect of all bits $i+1, \ldots, n$ on the function value. Both will later be needed in our construction.

For two search points $x, y \in \{0,1\}^n$, we write $x \leq y$ if $x_i \leq y_i$ holds for all $1 \leq i \leq n$. We write $x < y$ if $x \leq y$ and $x \neq y$ hold. We call $f$ a *strictly monotone function* (usually called simply *monotone* in the following) if for all $x, y \in \{0,1\}^n$ with $x < y$ it holds that $f(x) < f(y)$. Observe that the above condition is equivalent to $f(x) < f(y)$ for all $x$ and $y$ such that $x$ and $y$ only differ in exactly one bit and this bit has value 1 in $y$. In other words, every mutation that only flips bits with value 0 strictly increases the function value. Clearly, the all-ones bit string $1^n$ is the unique global optimum for a monotone function.

For the (1+1) EA, the difficulty of monotone functions strongly depends on the mutation probability $p(n)$. We are interested in mutation probabilities $p(n) = c/n$ for some constant $c \in \mathbb{R}^+$. For constants $c < 1$ on average in a single mutation less than one bit flips. If this is a 1-bit we have $f(x) > f(\text{mut}(x))$ and $x = x^+$ holds. Otherwise, $f(x^+) > f(x)$ holds and we accept this move. This way the number of 0-bits is quickly reduced to 0 and the unique global optimum is found. Using drift analysis this reasoning can easily be made precise. We state the result here and omit the proof due to space restrictions.

**Theorem 2.** *The expected optimization time of the (1+1) EA with mutation probability $p(n) = c/n$, $0 < c < 1$ constant, is $\Theta(n \log n)$ for every monotone function.*

The proof of Theorem 2 breaks down for $c = 1$. In this case the drift in the number of 1-bits can be bounded pessimistically by a model due to Jansen [Jan07] where we consider a random process that mutates $x$ to $y$ with mutation probability $p(n) = 1/n$ and replaces $x$ by $y$ if either $x \leq y$ holds or we have neither $x \leq y$ nor $y \leq x$ but $|y|_1 < |x|_1$ holds. This worst case model yields an upper bound of $O(n^{3/2})$ for the expected optimization time of the (1+1) EA with mutation probability $p(n) = 1/n$ on monotone functions.

**Theorem 3.** *The expected optimization time of the (1+1) EA with mutation probability $p(n) = 1/n$ is $O(n^{3/2})$ for every monotone function.*

Our main result is that using mutation probability $p(n) = c/n$ where $c$ is a sufficiently large constant, optimization of monotone functions can become very difficult for the (1+1) EA. This is the first result where increasing the mutation probability by a constant factor increases the optimization time from polynomial to exponential with overwhelming probability.

**Theorem 4.** *There exists a monotone function $f: \{0,1\}^n \to \mathbb{N}$ such that the (1+1) EA with mutation probability $p(n) = c/n$, $c \geq 33$ constant, does not optimize $f$ within $2^{\Omega(n)}$ mutations with overwhelming probability.*

The formal proof of this result is somewhat technical and lengthy. Therefore, in this extended abstract, we present how to construct such a monotone function $f$ in the following section. In Section 4, we describe why this function is difficult to optimize. Complete proofs are available from the authors by request.

## 3  A Difficult to Optimize Monotone Function

In this section, we describe a monotone function that is difficult to optimize via a (1+1) EA with mutation probability $p(n) = c/n$, if $c > 33$ is constant.

The main idea is the construction of a kind of long path function (compare the work by Horn, Goldberg, and Deb [HGD94]). We also have an exponentially long path such that shortcuts can only be taken if a large number of bits flip simultaneously, a very unlikely event. The construction is complicated by the fact that the function needs to be monotone. Hence we cannot forbid leaving the path by giving the boundary of the path an unfavorable fitness. We solve this problem, roughly speaking, by implementing the path on a level of bit strings having similar numbers of 1-bits. Monotonicity simply forbids leaving the level to strings having fewer 1-bits. The crucial part of our construction is setting up the function in such a way that, in spite of monotonicity, not too many 1-bits are collected.

For $x \in \{0,1\}^n$ let $B \subseteq [n]$ be a subset of all indices $[n]$. The bits $x_i$ with $i \in B$ are referred to as window. The bits $x_i$ with $i \notin B$ are outside the window. Inside the window the function value is given by BINVAL. The weights for BINVAL are ordered differently for each window in order to avoid correlation between windows. The window is placed such that there is only a small number of 0-bits outside the window. Reducing the number of 0-bits outside causes the window to be moved. This is a likely event that happens frequently. However, we manage to construct an exponentially long sequence of windows with the additional property that in order to come from one window to one with large distance in this sequence a large number of bits needs to be flipped simultaneously. Since this is highly unlikely, it is very likely that the sequence of windows is followed. This takes an exponential number of steps with overwhelming probability. Droste, Jansen, and Wegener [DJW98] embed the long path into a unimodal function in a way that the (1+1) EA reaches the beginning of the path with probability close to 1. We adopt this technique and extend it to our monotone function.

The following Lemma 5 defines the sequence of windows of our function by defining the index sets $B_i$. The property that windows with large distance have large Hamming distance is formally stated as $|i - j| \geq \ell \Rightarrow |B_i \cap B_j| \leq \gamma \ell$ for $\ell = \Theta(n)$ and some constant $\gamma > 0$.

**Lemma 5.** *Let $\beta, \gamma \in \mathbb{R}$ be constants with $0 < \beta$ and $\rho := \beta/(1-2\beta) < \gamma < 1$. Let $n \in \mathbb{N}$ and $L := \lfloor \exp((\gamma - \rho)^2 (1 - 2\beta)n/6) \rfloor$. Let $\ell := \beta n$ and $L' := L - \ell + 1$. Then there are $b_1, b_2, \ldots, b_L \in [n]$ such that the following holds. Let $B_i := \{b_i, b_{i+1}, \ldots, b_{i+\ell-1}\}$ for all $i \in [L']$. Then*

*(i) $|B_i| = \ell$ for all $i \in [L']$,*

*(ii) $|B_i \cap B_j| \leq \gamma \ell$ for all $i, j \in [L']$ such that $|i - j| \geq \ell$.*

*Proof.* The proof invokes the probabilistic method [AS00], that is, we describe a way to randomly choose the $b_i$ that ensures that properties (i) and (ii) hold with positive probability. This necessarily implies the existence of such a sequence.

Let the $b_1, b_2, \ldots, b_L$ be chosen uniformly at random subject to condition (i). More precisely, let $b_1 \in [n]$ be chosen uniformly at random. If $b_1, \ldots, b_{i-1}$ are already chosen, then choose $b_i$ from $[n] \setminus \{b_{\max\{1, i-\ell\}}, \ldots, b_{i-1}\}$ uniformly at random.

Let $i, j \in [L']$ with $i < j$ and $|i - j| \geq \ell$. By definition, the sets $B_i$ and $B_j$ do not share an index. Fix any outcome of $B_i$. For all $k \in \{0, \ldots, \ell - 1\}$ we have that, conditional on any outcomes of all other $b$s, the probability that $b_{j+k} \in B_i$ is at most $|B_i|/(n - 2\ell) = \beta/(1 - 2\beta)$. Consequently, the random variable $C = |B_i \cap B_j|$ is dominated by a random variable $X$ that is the sum of $\ell$ independent indicator random variables that are one with probability $\rho = \beta/(1-2\beta)$. Hence a simple Chernoff bound (cf. e.g. [MU05]) yields

$$\Pr(C > \gamma \ell) \leq \Pr(X > \gamma \ell) = \Pr(X > (1 + \tfrac{\gamma - \rho}{\rho}) \cdot \rho \ell) \leq \exp(-(\tfrac{\gamma - \rho}{\rho})^2 \rho \ell / 3).$$

Since there are less than $L^2$ choices of $i, j \in [L']$, a simple union bound yields

$$\Pr(\exists i, j \in [L'] \colon (|i - j| \geq \ell) \wedge (|B_i \cap B_j| > \gamma \ell)) < L^2 \exp(-(\tfrac{\gamma - \rho}{\rho})^2 \rho \ell / 3) \leq 1. \qquad \square$$

The following definition introduces the difficult monotone function we consider. Note that it assumes the sequence of windows $B_i$ to be given. For $x \in \{0,1\}^n$ some $i \in [L']$ is a *potential position* in the sequence of windows if the number of 0-bits outside the window $B_i$ is limited by $\alpha n$, $\alpha > 0$ some constant. We select the largest potential position $i$ as actual position and have the function value for $x$ depend mostly on this position. If no potential position $i$ exists, we have not yet found the path of windows and lead the (1+1) EA towards it. If $i = L'$, i.e., the end of the path is reached, the (1+1) EA is lead towards the unique global optimum via ONEMAX.

**Definition 6.** *Let $\beta$, $\gamma$, $\ell$, $L$, $L'$, the $b_i$ and $B_i$ be as in Lemma 5. Let $\alpha \in \mathbb{R}$ with $0 < \alpha < \beta$. For $x \in \{0,1\}^n$ let $\mathcal{B}_x := \{i \in [L'] \mid |\{j \in [n] \mid x_j = 0\} \setminus B_i| \leq \alpha n\}$. Let $i_x^* := \max \mathcal{B}_x$, if $\mathcal{B}_x$ is non-empty. For $i \in [L']$ let $\pi^{(i)}$ be a permutation of $B_i$. Denote by $\Pi = (\pi^{(1)}, \ldots, \pi^{(L')})$ the sequence of these permutations. We use the short-hand $\pi^{(i)}(x)$ to denote the vector obtained from permuting the components of $(x_{b_i}, \ldots, x_{b_{i+\ell-1}})$ according to $\pi^{(i)}$. Consequently, $\pi^{(i)}(x) = (x_{\pi^{(i)}(b_i)}, \ldots, x_{\pi^{(i)}(b_{i+\ell-1})})$.*

*We define $f_\Pi \colon \{0,1\}^n \to \mathbb{N}_0$ via*

$$f_\Pi(x) := \begin{cases} \left|x_{|[n]\setminus B_1}\right|_1 \cdot 2^n + \text{BinVal}(\pi^{(|x_{|[n]\setminus B_1}|_1)}(x)), & \text{if } \mathcal{B}_x = \emptyset, \\ i_x^* \cdot 2^{2n} + \text{BinVal}(\pi^{(n+i_x^*)}(x)), & \text{if } \mathcal{B}_x \neq \emptyset \text{ and } n + i_x^* < L', \\ L \cdot 2^{3n} + |x|_1, & \text{otherwise.} \end{cases}$$

We state one observation concerning the function $f_\Pi$ that is important in the following. It states that as long as the end of the path of windows is not found the number of 0-bits outside is not only bounded by $\alpha n$ but equals $\alpha n$. This property will be used later on to show that the window is moved frequently.

**Lemma 7.** *Let $f_\Pi \colon \{0,1\}^n \to \mathbb{N}_0$ be as in Definition 6. Let $x \in \{0,1\}^n$ with $\mathcal{B}_x \neq \emptyset$ and $i_x^* = \max \mathcal{B}_x$. If $n + i_x^* < L'$, then $|Z(x) \setminus B_{i_x^*}| = \alpha n$.*

*Proof.* By assumption we have $n + i_x^* < L'$. We consider $B_{n+i_x^*+1}$ and see that the set coincides with $B_{n+i_x^*}$ in all but two elements: we have $B_{n+i_x^*} \setminus B_{n+i_x^*+1} = \{b_{n+i_x^*}\}$ and $B_{i_x^*+1} \setminus B_{i_x^*} = \{b_{n+i_x^*+\ell}\}$. Consequently, $|Z(x) \setminus B_{n+i_x^*}|$ and $|Z(x) \setminus B_{n+i_x^*+1}|$ differ by at most one. Thus, $|Z(x) \setminus B_{n+i_x^*}| < \alpha n$ implies $|Z(x) \setminus B_{n+i_x^*+1}| \leq \alpha n$ and we can replace $i_x^*$ by $i_x^* + 1$. This contradicts $i_x^* = \max \mathcal{B}_x$. We have $|Z(x) \setminus B_{n+i_x^*}| \leq \alpha n$ by definition and thus $|Z(x) \setminus B_{n+i_x^*}| = \alpha n$ follows. □

Our first main claim is that $f_\Pi$ is in fact monotone. This is not difficult, but might, due to the complicated definition of $f_\Pi$, not be obvious.

**Lemma 8.** *For all $\Pi$ as above, $f_\Pi$ is monotone.*

*Proof.* Let $f := f_\Pi$. Let $x \in \{0,1\}^n$ and $j \in [n]$ such that $x_j = 0$. Let $y \in \{0,1\}^n$ be such that $y_k = x_k$ for all $k \in [n] \setminus \{j\}$ and $y_j = 1 - x_j$. That is, $y$ is obtained from $x$ by flipping the $j$th bit (which is zero in $x$) to one. To prove the lemma, it suffices to show $f(x) < f(y)$.

Let first $\mathcal{B}_x = \emptyset$. If $\mathcal{B}_y \neq \emptyset$ we have $f(x) < n \cdot 2^n + 2^n$ and $f(y) \geq 2^{2n}$ so $f(x) < f(y)$ follows. If $\mathcal{B}_y = \emptyset$ we have either $\left|x_{|[n]\setminus B_1}\right|_1 < \left|y_{|[n]\setminus B_1}\right|_1$ (in case $j \notin B_1$) or $\text{BinVal}(\pi^{(i)}(x)) < \text{BinVal}(\pi^{(i)}(y))$ (in case $j \in B_1$). In both cases, $f(x) < f(y)$ holds.

Now assume $\mathcal{B}_x \neq \emptyset$ and $n + i_x^* < L'$. By definition $\mathcal{B}_x \subseteq \mathcal{B}_y$, hence $i_y^* \geq i_x^*$. If $i_y^* = i_x^*$, then $j \in B_{i_x^*}$ and $f(y) > f(x)$ follows from $\text{BinVal}(\pi^{(n+i_y^*)}(y)) = \text{BinVal}(\pi^{(n+i_x^*)}(y)) > \text{BinVal}(\pi^{(n+i_x^*)}(x))$. If $i_y^* > i_x^*$, then $f(y) > f(x)$. In all other cases, $f(x) = L2^{3n} + |x|_1$ and $f(y) = L2^{3n} + |y|_1$, hence $f(y) > f(x)$. □

## 4 Proof of the Lower Bound

**Theorem 9.** *Consider the (1+1) EA with mutation probability $c/n$ for $c \geq 33$ on the function $f := f_\Pi$ from Definition 6 where $\Pi$ is chosen uniformly at random and the parameters are chosen according to $\beta := 10/131$, $\gamma := 20/221$, and $\alpha := 1/(1000c)$. There is a constant $\kappa > 0$ such that with probability $1 - 2^{-\Omega(n)}$ the (1+1) EA needs at least $2^{\kappa n}$ generations to optimize $f$.*

This result above shows that if $f$ is chosen randomly (according to the construction described), then the (1+1) EA w. o. p. needs an exponential time to find the optimum. Clearly, this implies that there exists a particular function $f$, that is, a choice of $\Pi$, such that the EA faces these difficulties. This is Theorem 4.

The proof for Theorem 9 is long and technical. Therefore, we only discuss the main proof ideas here. A complete proof can be found in the appendix.

Both after a typical initialization, when $\mathcal{B}_x = \emptyset$, and afterwards, when $\mathcal{B}_x \neq \emptyset$ and $n + i_x^* < L'$, we have the following situation. There is a window of bits ($B_{i_x^*}$ if $i_x^*$ is defined and $B_1$ otherwise) such that the fitness increases with BinVal as a function on the bits inside the window. Moreover, the fitness is always increased in case the mutation decreases the number of 0-bits outside the window. If $\mathcal{B}_x = \emptyset$ this is due to the term $\left|x_{|[n]\setminus B_1}\right|_1 \cdot 2^n$ in the fitness function and otherwise it is because the current $i_x^*$-value has increased. The gain in fitness is so large that it dominates any change of the bits in the window.

We claim that with this construction it is very likely that the current window always contains at least $\beta n/11$ 0-bits. This is proven by showing that in case the number of 0-bits in the window is in the interval $[\beta n/11, \beta n/10]$ then there is a tendency ("drift") to increase the number of 0-bits again. Applying a drift theorem by Oliveto and Witt [OW10] yields that even in an exponential number of generations the probability that the number of 0-bits in the window decreases below $\beta n/11$ is exponentially small. We first elaborate on why this drift holds and then explain how the lower bound of $\beta n/11$ 0-bits implies the claim.

If a mutation decreases the number of 0-bits outside the window, the bits inside the window are subject to random, unbiased mutations. Hence, if the number of 0-bits is at most $\beta n/10$ the expected number of bits flipping from 1 to 0 is larger than the expected number of bits flipping from 0 to 1. If the mutation probability is large enough, this makes up for the 0-bits lost outside the window and it leads to a net gain in 0-bits in expectation, with regard to the whole bit string. Note that the window is moved during such a mutation. As by Lemma 7 the number of 0-bits outside the window is fixed to $\alpha n$, we have a net gain in 0-bits for the window, regardless of its new position.

In case the number of 0-bits outside the window remains put, acceptance depends on a BinVal instance on the bits inside the window. For BinVal accepting the result of a mutation is completely determined by the flipping bit with the largest weight. In an accepted step, this bit must have flipped from 0 to 1. All bits with smaller weights have no impact on acceptance and therefore are subject to random, unbiased mutations. If, among all bits with smaller weights, there is a sufficiently small rate of 0-bits, more bits will flip from 1 to 0 than from 0 to 1. In this case, we again obtain a net increase in the number of 0-bits in the window, in expectation. Here we also require a large mutation probability since every increase of BinVal implies that one 0-bit has been lost and a surplus of flipping 1-bits has to make up for this loss. This holds in particular since the window only contains $\beta n$ bits and the surplus' absolute value must still be large.

For a fixed BinVal instance the bits tend to develop correlations between bit values and weights over time; bits with large weights are more likely to become 1 than bits with small weights. This development is disadvantageous since the above argument relies on many 1-bits with small weights. In order to break up these correlations we use random instances of BinVal wherever possible. These random instances change quickly. If $\mathcal{B}_x = \emptyset$ and, by Lemma 7, also if $n + i_x^* < L'$ we have at least $\alpha n$ 0-bits outside the current window and every mutation that flips exactly one of these bits leads to a new BinVal instance. Since this happens with probability $\Omega(1)$, this frequently prevents the algorithm from gathering 1-bits at bits with large weights. Pessimistically dealing with bits that have been touched by mutation while optimizing the same BinVal instance, a positive expected increase in the number of 0-bits can be shown.

How does the lower bound of $\beta n/11$ 0-bits inside the window imply Theorem 9? With overwhelming probability we start with $\mathcal{B}_x = \emptyset$ and at least $\beta n/10$ 0-bits in the window $B_1$. We maintain at least $\beta n/11$ 0-bits in $B_1$, while the algorithm is encouraged to turn the 0-bits outside of $B_1$ to 1 quickly. Once the number of 0-bits outside of $B_1$ has decreased to or below $\alpha n$, the path has been reached. The 0-bits in $B_1$ thereby ensure that the initial $i_x^*$-value is at most $\beta n$. The reason is that every two sets $B_i, B_j$ with $|i - j| \geq \ell$ only intersect in at most $\gamma \beta n$ bits, so $\beta n/11$ 0-bits in $B_i$ imply at least $\beta n/11 - \gamma \beta n$ 0-bits outside of $B_j$. For $j$ to become the new window, however, at most $\alpha n$ 0-bits outside of $B_j$ are allowed. By choice of $\alpha$, $\beta$, and $\gamma$, moving from $B_1$ to $B_j$ requires a linear number of 0-bits in $B_1$ to flip to 1 if $j > \beta n$. The described mutation has probability $n^{-\Omega(n)}$. The last argument also shows that the probability of increasing $i_x^*$ by more than $\beta n$ in one generation is $n^{-\Omega(n)}$. Hence, with overwhelming probability in each generation the (1+1) EA only makes progress at most $\beta n$ on the path. As the path has exponential length, the claimed lower bound follows.

## 5 Conclusions

In this work, we analyzed how the (1+1) EA optimizes monotone functions. We showed that the optimum of any monotone function is found efficiently if the mutation probability is at most $1/n$. Surprisingly, once the mutation probability exceeds $33/n$, the situation drastically changes. In this case, there are monotone functions that with high probability are not optimized within exponential time.

This results indicates that, to a greater extent than expected, care has to be taken when choosing the mutation probability, even if restricting oneself to mutation probabilities $c^*/n$ with a constant $c^*$. Contrary to previous observations, e. g., for linear functions, it may well happen that constant factor changes in the mutation probability lead to more than constant factor changes in the efficiency.

Besides generally suggesting more research on the right mutation probability, this work leaves two particular problems open. (i) For the mutation probability $1/n$, give a sharp upper bound for the optimization time of monotone functions (this order of magnitude is between $\Omega(n \log n)$ and $O(n^{3/2})$). (ii) Determine the largest constant $c^*$ such that the expected optimization time of the (1+1) EA with mutation probability $p(n) = c^*/n$ is $n^{O(1)}$ on every monotone function. Currently, we only know that $1 < c^* < 33$ holds.

**Acknowledgments.**

The third author was supported by a postdoctoral fellowship from the German Academic Exchange Service. This material is based in part upon works supported by the Science Foundation Ireland under Grant No. 07/SK/I1205.

# A Appendix

This appendix contains the full formal proofs of those statements for which only an explanation of the proof idea was given in the main part of the paper.

## A.1 Proof of Theorem 2

*Proof of Theorem 2.* Initially, there are at least $\lfloor n/2 \rfloor$ 0-bits in $x$ with probability at least $1/2$. Considering the situation after $(n/2)\ln n$ mutations a simple and direct calculation reveals that with probability $\Omega(1)$ at least one of these bits is never flipped [DJW02]. This yields $\Omega(n \log n)$ as lower bound.

For the upper bound we employ drift analysis [HY01]. First, we consider a distance measure $d \colon \{0,1\}^n \to \mathbb{R}_0^+$ with $d(x) := |x|_0$. We have

$$\begin{aligned}
\mathrm{E}(d(x) - d(x^+) \mid x) &= \Pr(x \neq x^+) \cdot \mathrm{E}(d(x) - d(x^+) \mid x \wedge x \neq x^+) \\
&\quad + \Pr(x = x^+) \cdot E(d(x) - d(x^+) \mid x \wedge x = x^+) \\
&= \Pr(x \neq x^+) \cdot \mathrm{E}(d(x) - d(x^+) \mid x \wedge x \neq x^+) \\
&\geq \binom{|x|_0}{1} \cdot \frac{c}{n} \cdot \left(1 - \frac{c}{n}\right)^{n-1} \cdot \mathrm{E}(d(x) - d(x^+) \mid x \wedge x \neq x^+) \\
&\geq \frac{c|x|_0}{n \cdot e^c} \cdot \mathrm{E}(d(x) - d(x^+) \mid x \wedge x \neq x^+) \\
&\geq \frac{c|x|_0}{n \cdot e^c} \cdot \left(1 - (n-1) \cdot \frac{c}{n}\right) \geq \frac{c(1-c)|x|_0}{n \cdot e^c}
\end{aligned}$$

and will use this lower bound on $\mathrm{E}(d(x) - d(x^+) \mid x)$ later. Note that in the calculation above only in the inequality

$$\mathrm{E}(d(x) - d(x^+) \mid x \wedge x \neq x^+) \geq \left(1 - (n-1) \cdot \frac{c}{n}\right)$$

we make use of the fact the fitness function $f$ is monotone. The event $x \neq x^+$ implies that there is at least one bit that was mutated from a 0-bit into a 1-bit. This contributes 1 to the expected difference. The remaining $n-1$ bits contribute only $-c(n-1)/n$ since each flips with mutation probability $c/n$, only.

Now, we consider a different distance measure $d' \colon \{0,1\}^n \to \mathbb{R}_0^+$ with $d'(x) := H_{d(x)}$ where $H_k = \sum_{i=1}^k 1/i$ denotes the $k$th Harmonic number. It is easy to see that $H_k - H_\ell \geq (k-\ell)/k$ holds for any $k, \ell \in \mathbb{N}$. We estimate

$$\begin{aligned}
\mathrm{E}(d'(x) - d'(x^+) \mid x) &= \mathrm{E}(H_{d(x)} - H_{d(x+)} \mid x) \geq \mathrm{E}\left(\frac{d(x) - d(x^+)}{d(x)} \mid x\right) \\
&= \frac{\mathrm{E}(d(x) - d(x^+) \mid x)}{d(x)} = \frac{\mathrm{E}(d(x) - d(x^+) \mid x)}{|x|_0} \geq \frac{c(1-c)}{n \cdot e^c}
\end{aligned}$$

and obtain

$$(\ln(n) + 1) \cdot \frac{n \cdot e^c}{c(1-c)} = O(n \log n)$$

as upper bound on the expected optimization time by application of the drift theorem [HY01] since the initial distance $d'(x)$ is bounded by $\ln(n) + 1$. $\square$

## A.2 Proof of Theorem 9

We consider the (1+1) EA with mutation probability $c/n$ and say that the (1+1) EA is on *level* $i_x^*$ if $x$ is the current search point. We also speak of *phase* $i_x^*$ as the random time until the (1+1) EA increases its current level. Note that many phases can be empty. $B_{i_x^*}$ is called the *current window* of bits in situations where we are looking at a trajectory of these sets and want to emphasize that the bits we are considering might change over time.

The main observation for our analysis is that the current window typically contains at least $\beta n/11$ 0-bits. This property is maintained even during an exponential number of generations, with overwhelming probability. Under this condition, the probability of increasing the current level $i_x^*$ by a large value is very small. Intuitively speaking, the reason for this is that the sets $B_i$ only have a small intersection and many bits have to change in order to move from $B_{i_x^*}$ to some set $B_j$ with $j \gg i_x^*$. This is made precise in the following lemma.

**Lemma 10.** *Let $0 < \alpha < \beta < \gamma$ be constants such that $1/11 - \alpha/\beta > \gamma > \beta/(1 - 2\beta)$. Let $f_\Pi$, with respect to $\alpha, \beta,$ and $\gamma$, be constructed as in Definition 6, for arbitrary $\Pi$. Let $x$ be the current search point of the (1+1) EA with mutation probability $c/n$ for a some $c \in \mathbb{R}^+$ optimizing $f_\Pi$. Assume that $\mathcal{B}_x \neq \emptyset$ and $B_{i_x^*}$ contains at least $\beta n/11$ 0-bits. Then the probability that the (1+1) EA increases the level $i_x^*$ by more than $\beta n$ in one generation is at most $n^{-\Omega(n)}$.*

*Proof.* Recall that $|B_{i_x^*} \cap B_j| \leq \gamma \beta n$ for all $j \geq i_x^* + \beta n$. If $B_{i_x^*}$ contains more than $\alpha n + \gamma \beta n$ 0-bits then there are more than $\alpha n$ 0-bits outside of $B_j$. Hence $j \notin \mathcal{B}_x$.

A necessary condition for increasing $i_x^*$ to any value $j \geq i_x^* + \beta n$ is thus that one mutation decreases the number of 0-bits in $B_i$ to a value below or equal to $\alpha n + \gamma \beta n$. This is a decrease of at least $\beta n/11 - \alpha n - \gamma \beta n =: \kappa n$ bits for some constant $0 < \kappa < 1$. The probability of flipping at least $\kappa n$ bits simultaneously is at most $\binom{n}{\kappa n} \cdot (c/n)^{\kappa n} \leq c^{\kappa n} \cdot 1/(\kappa n)! = n^{-\Omega(n)}$. □

One conclusion from this lemma is that, with overwhelming probability, the (1+1) EA follows the path given by the sets $B_i$, as each phase increases the current level by at most $\beta n$. This will establish the claimed time bound. The rest of this section deals with the proof of the invariance property on the number of 0-bits in the current window. For this proof we make use of the following drift theorem by Oliveto and Witt [OW10].

**Theorem 11** (Simplified Drift Theorem [OW10]). *Let $X_t$, $t \geq 0$, be the random variables describing a Markov process over a finite state space $S \subseteq \mathbb{R}_0^+$ and denote $\Delta_t(i) := (X_{t+1} - X_t \mid X_t = i)$ for $i \in S$ and $t \geq 0$. Suppose there exist an interval $[a, b]$ in the state space, two constants $\delta, \varepsilon > 0$ and, possibly depending on $L := b - a$, a function $r(L)$ satisfying $1 \leq r(L) = o(L/\log(L))$ such that for all $t \geq 0$ the following two conditions hold:*

1. $\mathrm{E}(\Delta_t(i)) \geq \varepsilon$ *for* $a < i < b$,
2. $\Pr(\Delta_t(i) \leq -j) \leq \frac{r(L)}{(1+\delta)^j}$ *for* $i > a$ *and* $j \in \mathbb{N}_0$.

*Then there is a constant $\kappa > 0$ such that for $T^* := \min\{t \geq 0 : X_t \leq a \mid X_0 \geq b\}$ it holds $\Pr(T^* \leq 2^{\kappa L/r(L)}) = 2^{-\Omega(L/r(L))}$.*

A prerequisite for this theorem is that the number of 0-bits in the current window increases in expectation, when the number of 0-bits is in a certain interval. We choose the interval $[\beta n/11, \beta n/10.5]$, but establish lower bounds for the drift with respect to a larger interval $[\beta n/11, \beta n/10]$. The larger interval will be used later on when proving that the after initialization the (1+1) EA finds the start of the path.

The "drift" on the number of 0-bits will be bounded from below by positive constants in two cases: either the current level remains fixed in one generation or the current level is increased.

We start with the former case and give a lower bound for the number of 0-bits in the current window.

## A.3 Invariance for Non-Sliding Windows

Before we formulate the main statement of this subsection, let us introduce some shorthands. For any $x$ let $x_B := x_{|B_{i_x^*}}$ denote the substring of $x$ induced by $B_{i_x^*}$, i.e., the substring in the current window. Recall that $|x_B|_0$ denotes the number of 0-bits in the current window. That is, $|x_B|_0 = |\{j \in \{b_{i_x^*}, \ldots b_{i_x^* + \ell - 1}\} \mid x_j = 0\}|$. For readability purposes we write $|x_B^+|_0$ instead of $|(x^+)_B|_0$ for the number of 0-bits of $x^+$ in its window. Note that, in this subsection, we always deal with the case that $i_x^* = i_{x^+}^*$ anyway.

In the following we show that, whenever the number of 0-bits in the current window is in the interval $[\frac{\beta n}{11}, \frac{\beta n}{10}]$, we observe a drift towards more 0-bits. This is formalized in the following lemma.

**Lemma 12.** *Let $0 < \alpha < \beta < 1$ and $c > \frac{5}{2\beta}$ be constants. Let $n$ be sufficiently large and let $f$, with respect to $\alpha$ and $\beta$, be constructed as in Definition 6. Let $x$ be the current search point of the (1+1) EA maximizing $f$. Assume $|x_B|_0 \in [\frac{\beta n}{11}, \frac{\beta n}{10}]$. We denote by $A$ the event that the (1+1) EA maximizing $f$ and starting in $x$ does not leave the current level, i.e., $i_x^* = i_{x^+}^*$. Then, the following two statements hold.*

1. *For every constant $\varepsilon > 0$ the number of different bits that are flipped during phase $i_x^*$ is at most $2\varepsilon\beta c n$, with probability $1 - \exp(-\Omega(n))$.*

2. *For small enough $\varepsilon > 0$, assuming that the event from 1. holds, there exists a constant $\delta > 0$ such that the drift in the number of 0-bits $\mathrm{E}(|x_B^+|_0 - |x_B|_0 \mid A)$ is at least $\delta$.*

The proof of this lemma will heavily depend on the drift in the number of 0-bits induced by the random BinVal within the current window. In the proof of Lemma 12, we will have to deal with variable lengths of the considered bit string. Therefore, the following auxiliary lemma is formulated for a bandwidth of possible bit string lengths.

**Lemma 13.** *Let $0 < \beta < 1$, $0 \leq \varepsilon < \beta$, and $c > 2\left(\frac{4}{5}\beta - \varepsilon\right)^{-1}$ be constants. Consider the (1+1) EA with mutation probability $c/n$ maximizing a random BinVal on $u \geq \beta n - \varepsilon n$ bits as defined in Definition 6. Let $\tilde{x}$ denote the current search point. Assume $|\tilde{x}|_0 \in [\frac{u}{11}, \frac{\beta n}{10}]$. If the random assignment of the function weights is independent of the position of the 0-bits $Z$, there exists a constant $\tilde{\delta} > 0$ such that the drift in the number of 0-bits within the current window is at least $\tilde{\delta}$, i.e., $\mathrm{E}(|\tilde{x}^+|_0 - |\tilde{x}|_0) \geq \tilde{\delta}$.*

In order to prevent confusion, let us remark that the expectation is drawn both with respect to the random assignment of the function weights as well as with respect to the position of the 0-bits of $\tilde{x}$.

*Proof of Lemma 13.* Let $\beta$, $\varepsilon$, $c$, $u$, and $\tilde{x}$ be as above. As a first observation, let us recall the following. Whenever $\tilde{x}^+ = \tilde{x}$, it holds that $|\tilde{x}^+|_0 - |\tilde{x}|_0 = 0$. Thus, we are only interested in the case $\tilde{x}^+ \neq \tilde{x}$. Note that in this case, the construction of BinVal implies that the bit with the largest weight is one that flips from 0 to 1 as the (1+1) EA would otherwise not accept $\mathrm{mut}(\tilde{x})$ as a new search point. For all other bits that are being flipped in this iteration, the direction of the flip bit (i.e., whether the bit itself is a 0-bit flipping to 1 or a 1-bit flipping to 0) is random and does only depend on the shares of 0- and 1-bits. This will be formalized in the following.

For readability purposes, let us introduce the following notations. For every $k \in \{0,\ldots,u\}$ we denote by $p_k$ the probability that the (1+1) EA flips exactly $k$ bits. Clearly, $p_k = \binom{u}{k}(\frac{c}{n})^k(1-\frac{c}{n})^{u-k}$ for $k \geq 1$ and $p_0 = (1-\frac{c}{n})^u$.

Let us, for the moment, assume that exactly $k$ bits are being flipped and let us consider the substring of the flipping bits only. If we remove from the substring the bit with the largest weight (which flips from 0 to 1), we get that the expected number of 0-bits in this reduced substring equals $(k-1)\frac{|\tilde{x}|_0-1}{u-1}$. Analogously, the expected number of 1-bits in the bit string equals $(k-1) - (k-1)\frac{|\tilde{x}|_0-1}{u-1}$. We thus obtain for this specific setting that the expected difference of $|\tilde{x}^+|_0 - |\tilde{x}|_0$ equals $\left((k-1) - (k-1)\frac{|\tilde{x}|_0-1}{u-1}\right) - (k-1)\frac{|\tilde{x}|_0-1}{u-1} - 1 = k(1-2\frac{|\tilde{x}|_0-1}{u-1}) - (2-2\frac{|\tilde{x}|_0-1}{u-1})$. Now, for any such $k$, it holds that $\mathrm{E}(|\tilde{x}^+|_0 - |\tilde{x}|_0 \mid k \text{ bits flip})$ equals the probability that the flipping bit with the largest weight flips from 0 to 1 (which occurs with probability $\frac{|\tilde{x}|_0}{u}$) times the drift conditional on $k$ bit flips and $\mathrm{mut}(\tilde{x}) \neq x$. The latter equals $k(1-2\frac{|\tilde{x}|_0-1}{u-1}) - (2-2\frac{|\tilde{x}|_0-1}{u-1})$ as outlined above.

Combining these observations, we gain the following.

$$\mathrm{E}(|\tilde{x}^+|_0 - |\tilde{x}|_0) = \sum_{k=1}^{u} p_k \frac{|\tilde{x}|_0}{u}\left[k(1-2\frac{|\tilde{x}|_0-1}{u-1}) - (2-2\frac{|\tilde{x}|_0-1}{u-1})\right].$$

Clearly, $\sum_{k=0}^{u} p_k = 1$ as we are dealing with a distribution. Thus, $\sum_{k=1}^{u} p_k = 1 - (1-\frac{c}{n})^u$. On the other hand, $\sum_{k=1}^{u} p_k k = \frac{c}{n}u \geq c(\beta-\varepsilon)$ as this sum equals the expected number of bit flips.

This yields

$$\mathrm{E}(|\tilde{x}^+|_0 - |\tilde{x}|_0) \geq \frac{|\tilde{x}|_0}{u}\left[(1-2\frac{|\tilde{x}|_0-1}{u-1})c(\beta-\varepsilon) - (2-2\frac{|\tilde{x}|_0-1}{u-1})(1-(1-\frac{c}{n})^u)\right].$$

Plugging in $\frac{|\tilde{x}|_0}{u} \geq \frac{1}{11}$ and $\frac{|\tilde{x}|_0-1}{u-1} \leq \frac{|\tilde{x}|_0}{u} \leq \frac{\beta n}{10u} \leq \frac{\beta}{10(\beta-\varepsilon)}$ yields

$$\mathrm{E}(|\tilde{x}^+|_0 - |\tilde{x}|_0) \geq \tfrac{1}{11}\left[(1-\tfrac{\beta}{5(\beta-\varepsilon)})c(\beta-\varepsilon) - 2\right] \geq \tfrac{1}{11}\left[c(\tfrac{4}{5}\beta-\varepsilon) - 2\right],$$

which, for $c > 2\left(\tfrac{4}{5}\beta-\varepsilon\right)^{-1}$, can clearly be bounded from below by some positive constant $\tilde{\delta}$. □

We can now easily deduct Lemma 12.

*Proof of Lemma 12.* Let $A$, $\beta$, $\alpha$, $f$, and $x$ be as in the statement of the lemma. Furthermore, let us assume that event $A$ holds. That is, the acceptance of the mutated bit string $\mathrm{mut}(x)$ is fully determined by the random BinVal within the current window. Thus, we can restrict our attention to the current window.

Let us begin with proving the first claim. For this purpose, let $\varepsilon > 0$ be a constant. We prove an auxiliary claim stating that with probability $\exp(-\Omega(n))$ the time $T_{i_x^*}$ until the (1+1) EA exits level $i_x^*$ is at most $\varepsilon n$. That is, we can assume that phase $i_{x^*}$ does not take longer than $\varepsilon n$ steps. We then show how to derive the original claim.

By construction, the (1+1) EA exits level $i_x^*$ if exactly one of the $\alpha n$ 0-bits outside the current window is being flipped. Thus, the probability $\Pr(i_x^* \neq i_{x+}^*)$ to exit the current level in one step is at least $\alpha n \cdot \frac{c}{n} \cdot (1-\frac{c}{n})^{n-1} \geq \alpha c e^{-c}(1-\frac{c}{n})^{c-1}$. It follows that the probability of not exiting level $i_x^*$ in $\varepsilon n$ steps is at most $(1-\alpha c e^{-c}(1-\frac{c}{n})^{c-1})^{\varepsilon n} \leq \exp(-\alpha c e^{-c}(1-\frac{c}{n})^{c-1}\varepsilon n) = \exp(-\Omega(n))$.

Now, the expected number of bits that have been flipped in $\varepsilon n$ steps is at most $\varepsilon n \cdot \beta n \cdot \frac{c}{n} = \varepsilon\beta cn$. We apply Chernoff bounds, and obtain that the probability that more than $2\varepsilon\beta cn$ bits are being flipped in $\varepsilon n$ steps is at most $\exp(-\tfrac{1}{3}\varepsilon\beta cn) = \exp(-\Omega(n))$.

We continue with the second claim. Therefore, let us assume that no more than $2\varepsilon\beta c n$ bits are being flipped during phase $i_x^*$. We note already here that we can conclude the following. The probability of flipping in the current iteration a bit that has already been flipped in a former iteration of phase $i_x^*$ is at most $2\varepsilon\beta c n \cdot \frac{c}{n} = 2\varepsilon\beta c^2$. That is, $\Pr(G \mid A) \leq 2\varepsilon\beta c^2$.

We denote by $G$ the event that in the current iteration, the (1+1) EA flips a bit that has already been flipped in a former iteration of phase $i_x^*$ and rewrite

$$\mathrm{E}(|x_B^+|_0 - |x_B|_0 \mid A) = \Pr(G \mid A)\,\mathrm{E}(|x_B^+|_0 - |x_B|_0 \mid A \wedge G)$$
$$+ (1 - \Pr(G \mid A))\,\mathrm{E}(|x_B^+|_0 - |x_B|_0 \mid A \wedge \bar{G}),$$

with $\bar{G}$ denoting the complementary event of $G$. Now, whenever $G$ occurs, we adopt a worst case view by assuming that all bits flip the wrong direction, i.e., from 0 to 1. For this purpose, let us, for the moment, assume that $G$ holds. In this case, at least one bit flips and we can, very pessimistically, assume that each of the flipping bits reduces the number of 0-bits by 1. Note that, given that one bit flips, the expected number of total bit flips in the current window equals $1 + \frac{c}{n}(\beta n - 1) < 1 + c\beta$. Thus, we can bound $\mathrm{E}(|x_B^+|_0 - |x_B|_0 \mid A \wedge G)$ from below by $-1 - c\beta$. That is, under our assumption, it holds that

$$\Pr(G \mid A)\,\mathrm{E}(|x_B^+|_0 - |x_B|_0 \mid A \wedge G) \geq 2\varepsilon\beta c^2(-1 - c\beta)\,.$$

We now need to give bounds for the second summand. For this purpose, we apply Lemma 13. As we are conditioning on $\bar{G}$, we apply the auxiliary lemma with $u$ denoting the number of bits that have not been flipped in any former iteration of phase $i_x^*$. Furthermore, we are only interested in the substring $\tilde{x}$ of $x$ consisting of these $u$ yet unflipped bits. As we have seen in the first part of this proof, with probability $1 - \exp(-\Omega(n))$ it holds that $u \geq \beta n - 2\varepsilon\beta c n$. As $\beta$ and $c$ are constants, we can choose $\varepsilon$ small enough, such that $2\varepsilon\beta c < \beta$. Furthermore, as $c > \frac{5}{2\beta}$, we can choose $\varepsilon$ small enough such that $c > 2\left(\frac{4}{5}\beta - 2\varepsilon\beta c\right)^{-1}$. Application of Lemma 13 yields $\mathrm{E}(|x_B^+|_0 - |x_B|_0 \mid A \wedge \bar{G}) \geq \tilde{\delta}$ for some positive constant $\tilde{\delta}$.

Altogether we obtain that,

$$\mathrm{E}(|x_B^+|_0 - |x_B|_0 \mid A) \geq 2\varepsilon\beta c^2(-1 - c\beta) + (1 - 2\varepsilon\beta c^2)\tilde{\delta}\,.$$

Lastly, we observe that we can choose $\varepsilon$ small enough such that this term can be bounded from below by some positive constant $\delta$, as claimed. $\square$

### A.4 Invariance for Sliding Windows

We now consider the case where the current level is increased, i.e. a transition from $i_x^*$ to $i_{x^+}^*$ with $i_x^* < i_{x^+}^* < L'$ happens. Recall that $x_B := x_{|B_{i_x^*}}$ denotes the substring of $x$ induced by $B_{i_x^*}$, i.e., the substring in the current window. Moreover, let $x_B^+ := x_{|B_{i_{x^+}^*}}^+$ denote the substring of $x^+$ induced by $B_{i_{x^+}^*}$. Note that, in this subsection, we deal with the case $i_x^* \neq i_{x^+}^*$ and thus, $B_{i_x^*} \neq B_{i_{x^+}^*}$ and $x_B^+ \neq x_B$ holds. As before, let $|x_B|_0$ denote the number of 0-bits of $x$ and $|x_B^+|_0$ the number of 0-bits of $x^+$ in the corresponding current window.

We show that also in this situation we have a drift in the number of 0-bits within the current window that is bounded below by a positive constant. Due to the transition it is no longer sufficient to only consider changes within the current window. Furthermore, transitions are often triggered by changes outside the current window. Thus, we switch to a form of a "global" view and take into account both the changes within the current window as well as changes outside the current window. We formalize this within the next lemma.

**Lemma 14.** *Let $0 < \alpha < \beta < 1$ with $5/(2\beta) < c < 1/(3\alpha)$ be constants. Let $n$ be sufficiently large and let $f$, with respect to $\alpha$ and $\beta$, be constructed as in Theorem 9. Let $x$ be the current search point of the (1+1) EA maximizing $f$. We denote by $\bar{A}$ the event that a transition from level $i_x^*$ to $i_{x^+}^*$ with $i_x^* < i_{x^+}^* < L'$ occurs in an iteration of the (1+1) EA maximizing $f$. Assume $|x_B|_0 \leq (1/10)\beta n$.*

*Then, there is a constant $\delta > 0$ such that the drift in the number of 0-bits is at least $\delta$, i.e., $\mathrm{E}(|x_B^+|_0 - |x_B|_0 \mid \bar{A}) \geq \delta$.*

*Proof.* Let $\overline{B}_{i_x^*} = [n] \setminus B_{i_x^*}$, the indices not contained in the current window, and $x_{\overline{B}} := x_{|\overline{B}_{i_x^*}}$ the corresponding induced substring of $x$. Analogously, we define $x_{\overline{B}}^+ := x_{|\overline{B}_{i_{x^+}^*}}$. Due to Lemma 7 we have $|x_{\overline{B}}|_0 = |x_{\overline{B}}^+|_0 = \alpha n$.

The main part of the proof is to derive a lower bound on $\mathrm{E}(|x_B^+|_0 \mid \bar{A})$. Afterwards we show that this bound together with the given prerequisites on $\alpha$, $\beta$, $c$ and $|x_B|_0$ yields a positive drift in the number of 0-bits.

It is easy to see that, conditional on $\bar{A}$, the expected number of 0-bits in the new window $B_{i_{x^+}^*}$ after a transition from $i_x^*$ to $i_{x^+}^*$ can be derived as the difference of the expected number of 0-bits in the current window $B_{i_x^*}$ after mutation and the expected amount of 0-bits lost outside the current window due to mutation:

$$\mathrm{E}\left(|x_B^+|_0 \mid \bar{A}\right) = \mathrm{E}\left(|\mathrm{mut}(x_B)|_0 \mid \bar{A}\right) - \mathrm{E}\left(|x_{\overline{B}}|_0 - |\mathrm{mut}(x_{\overline{B}})|_0 \mid \bar{A}\right) \tag{1}$$

We derive bounds for both parts of (1) separately. We start with an upper bound on the expected number of 0-bits in the current window after mutation, i.e., $\mathrm{E}(|\mathrm{mut}(x_B)|_0 \mid \bar{A})$ by the following case distinction.

In the first case, the transition happens independently of the change in the window. This case happens with probability $\Omega(1)$ as a 1-bit mutation of one of the $\alpha n$ 0-bits outside the current window suffices. In this situation, the expected number of 0-bits in the window is independent of $\bar{A}$ and thus, can be easily calculated as follows.

$$\begin{aligned}\mathrm{E}\left(|\mathrm{mut}(x_B)|_0\right) &= \left(1 - \frac{c}{n}\right) \cdot |x_B|_0 + \frac{c}{n} \cdot |x_B|_1 = |x_B|_0 - \frac{c}{n} \cdot |x_B|_0 + \frac{c}{n} \cdot |x_B|_1 \\ &= |x_B|_0 + \frac{c}{n} \cdot (\beta n - 2|x_B|_0) = |x_B|_0 + c\beta - \frac{2c|x_B|_0}{n}\end{aligned}$$

For the second case, i.e., if the mutation within the current window has influence on the transition performed, we have to be more careful as the expected number of 0-bits within the window is no longer independent of $\bar{A}$. However, before the mutation the leftmost bit in the current window is zero since otherwise we would have been able to increase the fitness by performing a transition by exactly one position. If this leftmost zero is flipped a transition is performed. Furthermore, it is necessary to flip this leftmost zero if the mutation within the window is supposed to have influence on the transition performed. The probability to flip this bit is $c/n$ and thus, the probability for this case is at most $c/n$.

We bound the contribution of this case in a pessimistic way. Similar to Lemma 10 we see that the number of bits flipping in one single iteration is at most $O(\log n)$ with probability $1 - n^{-\omega(1)}$. Furthermore, the contribution is at most $\beta n$ otherwise. Altogether, this yields a contribution to the expected value of at most

$$\frac{c}{n} \cdot \left(\left(1 - n^{-\omega(1)}\right) \cdot \log n + n^{-\omega(1)} \cdot \beta n\right) = O\left(\frac{\log n}{n}\right)$$

and we get the following lower bound on the expected number of 0-bits within the current window after mutation.

$$\mathrm{E}\left(|\mathrm{mut}(x_B)|_0 \mid \bar{A}\right) \geq |x_B|_0 + c\beta - \frac{2c|x_B|_0}{n} - O\left(\frac{\log n}{n}\right) \tag{2}$$

The second part of (1), i. e. the expected loss of 0-bits outside the current window due to mutation, is more difficult. For the sake of readability, let $k := |x_{\bar{B}}|_0 - |\mathrm{mut}(x_{\bar{B}})|_0$, the loss of 0-bits outside the current window due to mutation. Then, we are searching for $\mathrm{E}(k \mid \bar{A})$. We distinguish two cases. If $k > 0$ we definitely observe a transition and accept the new search point. If $k \leq 0$ a transition does not necessarily occur. For $k < 0$ the new search point is only accepted if this is the case. For $k = 0$ the search might also be accepted depending on the changes within the current window. We see that $\mathrm{E}(k) \leq \mathrm{E}(k \mid \bar{A}) \leq \mathrm{E}(k \mid k > 0)$ holds.

Let $Z_0$ be the number of 0-bits flipping to one and $Z_1$ the number of 1-bits flipping to zero. This yields $\mathrm{E}(k \mid k > 0) = \mathrm{E}(Z_0 - Z_1 \mid Z_0 > Z_1)$. With $|x_{\bar{B}}|_0 = \alpha n$ and $|x_{\bar{B}}|_1 = (1 - \alpha - \beta)n$ we get the following transformation of the expected value sought.

$$
\begin{aligned}
&\mathrm{E}(Z_0 - Z_1 \mid Z_0 > Z_1) \\
=\ & \sum_{i=0}^{(1-\alpha-\beta)n} \Pr(Z_1 = i) \cdot \mathrm{E}(Z_0 - Z_1 \mid Z_0 > Z_1 \text{ and } Z_1 = i) \\
=\ & \sum_{i=0}^{(1-\alpha-\beta)n} \Pr(Z_1 = i) \cdot \left(\mathrm{E}(Z_0 \mid Z_0 > i) - \mathrm{E}(Z_1 \mid Z_0 > Z_1 \text{ and } Z_1 = i)\right) \\
=\ & \sum_{i=0}^{(1-\alpha-\beta)n} \Pr(Z_1 = i) \cdot \left(\mathrm{E}(Z_0 \mid Z_0 > i) - i\right) \\
=\ & \sum_{i=0}^{(1-\alpha-\beta)n} \Pr(Z_1 = i) \cdot \left(\sum_{j=0}^{\alpha n} j \cdot \Pr(Z_0 = j \mid Z_0 > i) - i\right) \\
=\ & \sum_{i=0}^{(1-\alpha-\beta)n} \Pr(Z_1 = i) \cdot \left(\sum_{j=0}^{\alpha n} j \cdot \frac{\Pr(Z_0 = j \text{ and } Z_0 > i)}{\Pr(Z_0 > i)} - i\right) \\
=\ & \sum_{i=0}^{(1-\alpha-\beta)n} \Pr(Z_1 = i) \cdot \left(\frac{\sum_{j=i+1}^{\alpha n} j \cdot \Pr(Z_0 = j)}{\Pr(Z_0 > i)} - i\right) \\
\stackrel{(*)}{=}\ & \sum_{i=0}^{(1-\alpha-\beta)n} \Pr(Z_1 = i) \cdot \left(\frac{(i+1)\Pr(Z_0 > i) + \sum_{j=i+1}^{\alpha n} \Pr(Z_0 > j)}{\Pr(Z_0 > i)} - i\right) \\
=\ & \sum_{i=0}^{(1-\alpha-\beta)n} \Pr(Z_1 = i) \cdot \left(1 + \frac{\sum_{j=i+1}^{\alpha n} \Pr(Z_0 > j)}{\Pr(Z_0 > i)}\right) \tag{3}
\end{aligned}
$$

Note that in $(*)$ we used the following transformation.

$$
\begin{aligned}
&\sum_{j=i+1}^{\alpha n} j \cdot \Pr(Z_0 = j) \\
={} & (i+1) \cdot \Pr(Z_0 = i+1) + (i+1) \cdot \Pr(Z_0 = i+2) + \cdots + (i+1) \cdot \Pr(Z_0 = \alpha n) \\
& \hspace{3.2cm} + \Pr(Z_0 = i+2) \hspace{1.6cm} + \cdots + \Pr(Z_0 = \alpha n) \\
& \hspace{5.6cm} + \Pr(Z_0 = i+3) + \cdots + \Pr(Z_0 = \alpha n) \\
& \hspace{7cm} \ddots \\
& \hspace{9.8cm} + \Pr(Z_0 = \alpha n) \\
={} & (i+1) \cdot \Pr(Z_0 > i) + \Pr(Z_0 > i+1) + \Pr(Z_0 > i+2) + \cdots + \Pr(Z_0 > \alpha n - 1) \\
={} & (i+1) \cdot \Pr(Z_0 > i) + \sum_{j=i+1}^{\alpha n} \Pr(Z_0 > j)
\end{aligned}
$$

It is easy to see the following estimations for the probabilities used above.

$$
\begin{aligned}
\Pr(Z_0 > j) &\le \binom{\alpha n}{j+1} \cdot \left(\frac{c}{n}\right)^{j+1} \\
\Pr(Z_0 > i) &\ge \binom{\alpha n}{i+1} \cdot \left(\frac{c}{n}\right)^{i+1} \cdot \left(1 - \frac{c}{n}\right)^{\alpha n - i - 1} \\
\Pr(Z_1 = i) &= \binom{(1-\alpha-\beta)n}{i} \cdot \left(\frac{c}{n}\right)^i \cdot \left(1 - \frac{c}{n}\right)^{(1-\alpha-\beta)n - i}
\end{aligned}
$$

Plugging these inequalities into (3) yields the following expression for the expected loss of 0-bits outside the current window due to mutation.

$$
\begin{aligned}
\mathrm{E}(k \mid \bar{A}) \le{} & \sum_{i=0}^{(1-\alpha-\beta)n} \binom{(1-\alpha-\beta)n}{i} \cdot \left(\frac{c}{n}\right)^i \cdot \left(1 - \frac{c}{n}\right)^{(1-\alpha-\beta)n - i} \\
& \cdot \left( 1 + \frac{\sum_{j=i+1}^{\alpha n} \binom{\alpha n}{j+1} \cdot \left(\frac{c}{n}\right)^{j+1}}{\binom{\alpha n}{i+1} \cdot \left(\frac{c}{n}\right)^{i+1} \cdot \left(1 - \frac{c}{n}\right)^{\alpha n - i - 1}} \right) \\
={} & 1 + \sum_{i=0}^{(1-\alpha-\beta)n} \binom{(1-\alpha-\beta)n}{i} \cdot \left(\frac{c}{n}\right)^i \cdot \left(1 - \frac{c}{n}\right)^{(1-\alpha-\beta)n - i} \\
& \cdot \frac{\sum_{j=i+1}^{\alpha n} \binom{\alpha n}{j+1} \cdot \left(\frac{c}{n}\right)^{j+1}}{\binom{\alpha n}{i+1} \cdot \left(\frac{c}{n}\right)^{i+1} \cdot \left(1 - \frac{c}{n}\right)^{\alpha n - i - 1}}
\end{aligned}
\quad (4)
$$

We start with the second part of this term and derive the following lower bound.

$$
\frac{\sum_{j=i+1}^{\alpha n} \binom{\alpha n}{j+1} \cdot \left(\frac{c}{n}\right)^{j+1}}{\binom{\alpha n}{i+1} \cdot \left(\frac{c}{n}\right)^{i+1} \cdot \left(1 - \frac{c}{n}\right)^{\alpha n - i - 1}}
$$

$$
\begin{aligned}
&= \frac{\sum_{j=i+1}^{\alpha n} \frac{(\alpha n)!}{(j+1)!(\alpha n-j-1)!} \cdot \left(\frac{c}{n}\right)^{j+1} \cdot \frac{(i+1)!(\alpha n-i-1)!}{(\alpha n)!} \cdot \left(\frac{n}{c}\right)^{i+1}}{\left(1-\frac{c}{n}\right)^{\alpha n-i-1}} \\
&= \frac{1}{\left(1-\frac{c}{n}\right)^{\alpha n-i-1}} \cdot \sum_{j=i+1}^{\alpha n} \left(\frac{c}{n}\right)^{j-i} \cdot \frac{(i+1)!}{(j+1)!} \cdot \frac{(\alpha n-i-1)!}{(\alpha n-j-1)!} \\
&= \frac{1}{\left(1-\frac{c}{n}\right)^{\alpha n-i-1}} \cdot \sum_{j=i+1}^{\alpha n} \left(\frac{c}{n}\right)^{j-i} \cdot \frac{\alpha n-i-1}{i+2} \cdot \frac{\alpha n-i-2}{i+3} \cdots \frac{\alpha n-j}{j+1} \\
&\leq \frac{1}{\left(1-\frac{c}{n}\right)^{\alpha n-i-1}} \cdot \sum_{j=i+1}^{\alpha n} \left(\frac{c}{n}\right)^{j-i} \cdot \left(\frac{\alpha n}{i+1}\right)^{j-i} \\
&= \frac{1}{\left(1-\frac{c}{n}\right)^{\alpha n-i-1}} \cdot \sum_{j=i+1}^{\alpha n} \left(\frac{c\alpha}{i+1}\right)^{j-i} \\
&= \frac{1}{\left(1-\frac{c}{n}\right)^{\alpha n-i-1}} \cdot \sum_{j=1}^{\alpha n-i} \left(\frac{c\alpha}{i+1}\right)^{j} \\
&\leq \frac{1}{\left(1-\frac{c}{n}\right)^{\alpha n-i-1}} \cdot \frac{\frac{c\alpha}{i+1}}{1-\frac{c\alpha}{i+1}} \\
&\leq \frac{1}{\left(1-\frac{c}{n}\right)^{\alpha n-i-1}} \cdot \frac{c\alpha}{1-c\alpha}
\end{aligned}
$$

Remember, that we assume $c < 1/(3\alpha)$ and thus, $\alpha c < 1$ holds. Therefore, we can further simplify the above inequality by using the following simple estimation.

$$
\left(1-\frac{c}{n}\right)^{\alpha n-i-1} \geq \left(1-\frac{c}{n}\right)^{\alpha n} = \left(1-\frac{c}{n}\right)^{\left(\frac{n}{c}-1\right)\alpha c} \cdot \left(1-\frac{c}{n}\right)^{\alpha c}
$$

$$
\geq e^{-\alpha c} \cdot \left(1-\frac{c}{n}\right)^{\alpha c} = \left(\frac{1-\frac{c}{n}}{e}\right)^{\alpha c}
$$

Plugging all this into (4) yields a lower bound on the expected value sought.

$$
\begin{aligned}
\mathrm{E}(k \mid \bar{A}) &\leq 1 + \sum_{i=0}^{(1-\alpha-\beta)n} \binom{(1-\alpha-\beta)n}{i} \cdot \left(\frac{c}{n}\right)^{i} \cdot \left(1-\frac{c}{n}\right)^{(1-\alpha-\beta)n-i} \\
&\quad \cdot \left(\frac{e}{1-\frac{c}{n}}\right)^{\alpha c} \cdot \frac{c\alpha}{1-c\alpha} \\
&= 1 + \left(\frac{e}{1-\frac{c}{n}}\right)^{\alpha c} \cdot \frac{c\alpha}{1-c\alpha} \quad (5)
\end{aligned}
$$

We are now able to put the results from (2) and (5) together to get a lower bound on (1).

$$
\begin{aligned}
\mathrm{E}\left(\left|x_{B}^{+}\right|_{0} \mid \bar{A}\right) &= \mathrm{E}\left(|\mathrm{mut}(x_{B})|_{0} \mid \bar{A}\right) - \mathrm{E}\left(|x_{\overline{B}}|_{0} - |\mathrm{mut}(x_{\overline{B}})|_{0} \mid \bar{A}\right) \\
&\geq |x_{B}|_{0} + c\beta - \frac{2c|x_{B}|_{0}}{n} - O\left(\frac{\log n}{n}\right) - \left(1 + \left(\frac{e}{1-\frac{c}{n}}\right)^{\alpha c} \cdot \frac{c\alpha}{1-c\alpha}\right)
\end{aligned}
$$

With $|x_{B}|_{0} \leq (1/10)\beta n$, $5/(2\beta) < c < 1/(3\alpha)$ this yields the following lower bound on the drift in the number of 0-bits.

$$\begin{aligned}
\mathrm{E}(|x_B^+|_0 - |x_B|_0 \mid \bar{A}) &\geq c\beta - \frac{2c\,|x_B|_0}{n} - O\left(\frac{\log n}{n}\right) - \left(1 + \left(\frac{e}{1-\frac{c}{n}}\right)^{\alpha c} \cdot \frac{c\alpha}{1-c\alpha}\right) \\
&\geq c\beta - \frac{2c \cdot \frac{1}{10}\beta n}{n} - O\left(\frac{\log n}{n}\right) - 1 - \left(\frac{e}{1-\frac{c}{n}}\right)^{\frac{1}{3}} \cdot \frac{\frac{1}{3}}{1-\frac{1}{3}} \\
&= \frac{4c\beta}{5} - 1 - \frac{\sqrt[3]{e}}{2}\sqrt[3]{\frac{n}{n-c}} - O\left(\frac{\log n}{n}\right) \\
&\geq \frac{4c \cdot \frac{5}{2c}}{5} - 1 - \frac{\sqrt[3]{e}}{2}\sqrt[3]{\frac{n}{n-c}} - O\left(\frac{\log n}{n}\right) \\
&= 1 - \frac{\sqrt[3]{e}}{2}\sqrt[3]{\frac{n}{n-c}} - O\left(\frac{\log n}{n}\right)
\end{aligned}$$

Since $\sqrt[3]{e}/2 < 1$, $\lim_{n\to\infty} \sqrt[3]{n/(n-c)} = 1$ and $\lim_{n\to\infty} \log n/n = 0$ this is bounded below by a positive constant $\delta$ for sufficiently large values of $n$ which concludes the proof. □

Finally, we are able to prove the claimed invariance property.

**Lemma 15.** *Let $0 < \alpha < \beta < 1$ be constants such that $\gamma > \beta/(1-2\beta)$, $c > 5/(2\beta)$, and $\alpha > 1/(3c)$. Let $f_\Pi$, with respect to $\alpha$ and $\beta$, be constructed as in Definition 6, for $\Pi$ chosen uniformly at random.*

*Assume that for the current search point $x$ of the (1+1) EA it holds $\mathcal{B}_x \neq \emptyset$ and the current window contains at least $\beta n/10.5$ 0-bits. There is a constant $\kappa > 0$ such that with probability $1 - 2^{-\Omega(n)}$ in the following $2^{\kappa n}$ generations the (1+1) EA always has at least $\beta n/11$ 0-bits in the current window or the end of the path is reached.*

*Proof.* First observe that the event described in the first statement of Lemma 12 occurs with probability $1 - 2^{-\Omega(n)}$. By the union bound, the probability that the event occurs within $2^{\kappa n}$ phases is still $1 - 2^{-\Omega(n)}$ if $\kappa > 0$ is a sufficiently small constant.

We apply the drift theorem (Theorem 11) to a potential that reflects the number of 0-bits in the current window. Consider the interval $[\beta n/11, \beta n/10.5]$ and observe that by assumption the algorithm starts with a potential of at least $\beta n/10.5$. Using Lemma 12 with the condition from the first paragraph and Lemma 14, if the current potential is within the interval and the end of the path is not reached then the expected increase in the potential is bounded from below by a positive constant.

For $j \in \mathbb{N}_0$ the probability that the potential decreases by $j$ is bounded from above by the probability that the (1+1) EA flips at least $j$ bits. This probability is at most $\binom{n}{j}(c/n)^j \leq c^j/(j!) \leq (ec/j)^j \leq 2^{-j} \cdot 2^{2ec}$, where the last estimation is trivial for $j \leq 2ec$ and obvious otherwise. Applying Theorem 11 with $\delta := 1$ and $r := 2^{2ec}$ yields that with overwhelming probability in $2^{\kappa n}$ generations, if again $\kappa$ is sufficiently small, the potential does not decrease below $\beta n/11$ or the end of the path is reached. □

All that is left to complete the proof of the main result is the fact that the path is reached from a random initialization, with overwhelming probability.

**Lemma 16.** *Let $0 < \alpha < \beta < \gamma < 1$ be constants such that $1/11 - \alpha/\beta > \gamma > \beta/(1-2\beta)$ and $c > 5/(2\beta)$. Let $f_\Pi$, with respect to $\alpha, \beta$, and $\gamma$, be constructed as in Definition 6, for $\Pi$ chosen uniformly at random. With probability $1 - 2^{-\Omega(n)}$ the (1+1) EA with mutation probability $c/n$ for $c \geq 33$ optimizing $f_\Pi$ at some point of time reaches some search point $x$ with $i_x^* \leq \beta n$ and $|x_{|B_{i_x^*}}|_0 \geq \beta n/10.5$.*

*Proof.* Let $x$ be the first search point where $\mathcal{B}_x \neq \emptyset$. If $|x_{|B_1}|_0 \geq \beta n/11$ then for every $j > \beta n$, since $\beta n/11 > \alpha n + \gamma \beta n$ and $|B_1 \cap B_j| \leq \gamma \beta n$, we have $j \notin \mathcal{B}_x$. Hence, we only need to prove that the number of 0-bits in $B_1$ does not decrease below $\beta n/11$ until $\mathcal{B}_x$ becomes non-empty for the first time.

The set $\mathcal{B}_x$ is non-empty if the number of 0-bits outside of $B_1$ has decreased towards a value at most $\alpha n$. As every mutation decreasing the number of 0-bits outside of $B_1$ is accepted and such a mutation has a probability of at least $\alpha n \cdot c/n(1-c/n)^{n-1} = \Omega(1)$, for any initialization the expected number of generations until we have decreased the number of 0-bits to a value at most $\alpha n$ is $O(n)$. In addition, by Chernoff bounds the probability that more than $n^2$ generations are necessary is $2^{-\Omega(n)}$.

The probability that the initial search point contains at least $\beta n/10$ 0-bits in $B_1$ is $1 - 2^{-\Omega(n)}$ by Chernoff bounds. Assume that this happens and consider a situation where we have at least $\alpha n$ 0-bits outside of $B_1$. Arguing as in the proof of Lemma 12, if the number of 0-bit is within $\beta n/11$ and $\beta n/10$ then there is a positive drift. Instead of considering a new random BINVAL instance when the current $i_x^*$-value is increased, we obtain a new BINVAL instance whenever the number of 1-bits outside the window is increased. (The probability for the latter event can even be larger than the probability for the former.) This allows us to apply Lemma 13 in the same fashion as in the proof of Lemma 12. This results in a positive drift and applying the drift theorem as in Theorem 15 w.r.t. the interval $[\beta n/10.4, \beta n/10]$ proves that in $n^2$ generations the number of 0-bits in $B_1$ does not drop to or below $\beta n/10.4$, with probability $1 - 2^{-\Omega(n)}$.

Consider the mutation that creates $x$. Since $x$ is the first search point where $\mathcal{B}_x \neq \emptyset$, its parent must have had more than $\alpha n$ 0-bits outside of $B_1$. The probability that during mutation more than $\beta n/1100$ bits were flipped is $n^{-\Omega(n)}$. Hence $|x|_0 \geq \alpha n - \beta n/1100 + \beta n/10.4 \geq \alpha n + \beta n/10.5$. By Lemma 7 we then have $|x_{|B_{i_x^*}}|_0 = |x|_0 - \alpha n \geq \beta n/10.5$. As the sum of all error probabilities is $2^{-\Omega(n)}$, the claim follows. □

Now we are prepared to prove Theorem 9.

*Proof of Theorem 9.* It is easily verified that for the chosen values $0 < \alpha < \beta < \gamma < 1$, $c > 5/(2\beta)$, $1/11 - \alpha/\beta > \gamma > \beta/(1-2\beta)$, and $\alpha < 1/(3c)$ holds, satisfying all preconditions on these variables for Lemmas 10, 15, and 16. By Lemma 16 the (1+1) EA reaches some search point $x$ with $i_x^* \leq \beta n$ and $|x_{|B_{i_x^*}}|_0 \geq \beta n/10.5$ with overwhelming probability. Lemma 15 then states that with probability $1 - 2^{-\Omega(n)}$ the number of 0-bits in the current window is always at least $\beta n/11$ until the end of the path is reached or $2^{\kappa n}$ generations have passed for a sufficiently small constant $\kappa > 0$ (which would correspond to the claimed time bound).

Given the condition on the 0-bits, by Lemma 10 the (1+1) EA increases its current $i_x^*$-value by at most $\beta n$ in one generation, with probability $1 - n^{-\Omega(n)}$. The probability that this always happens until an $i_x^*$-value of $L'$ is reached is at least $1 - L' \cdot n^{-\Omega(n)} = 1 - n^{-\Omega(n)}$ since $L' = 2^{\Theta(n)}$. This implies that (1+1) EA spends at least $L'/(\beta n) - 1 \geq 2^{\kappa n}$ generations on the path, with probability $1 - 2^{-\Omega(n)}$, if $\kappa$ is chosen small enough. Since the sum of all error probabilities is $2^{-\Omega(n)}$, the claim follows. □